\pdfoutput=1

\documentclass[11pt]{article}

\usepackage{acl}

\usepackage{times}
\usepackage{latexsym}
\usepackage{graphicx}
\usepackage{tabularx}
\usepackage{subfigure}
\usepackage{amsmath}
\usepackage{xcolor}
\usepackage{amssymb}
\usepackage{array}
\usepackage{times}
\usepackage{latexsym}
\usepackage{multirow}
\usepackage{array}

\usepackage[T1]{fontenc}

\usepackage[utf8]{inputenc}

\usepackage{microtype}
\usepackage{booktabs}
\usepackage{multirow}
\usepackage{booktabs}
\usepackage{amssymb}
\usepackage{graphicx}
\usepackage{times}
\usepackage{latexsym}
\usepackage{url}
\usepackage[normalem]{ulem}
\useunder{\uline}{\ul}{}
\title{Empathetic Dialogue Generation via Sensitive Emotion Recognition and Sensible Knowledge Selection}

\author{Lanrui Wang\textsuperscript{\rm 1,2}, Jiangnan Li\textsuperscript{\rm 1,2}, Zheng Lin\textsuperscript{\rm 1,2}\thanks{\ \ \ Zheng Lin is the corresponding author. },  Fandong Meng\textsuperscript{\rm 3},  \\
{\bf Chenxu Yang\textsuperscript{\rm 1,2}, Weiping Wang\textsuperscript{\rm 1},  Jie Zhou\textsuperscript{\rm 3}}  \\
  \textsuperscript{\rm 1}Institute of Information Engineering, Chinese Academy of Sciences, Beijing, China \\
  \textsuperscript{\rm 2}School of Cyber Security, University of Chinese Academy of Sciences, Beijing, China \\
  \textsuperscript{\rm 3}Pattern Recognition Center, WeChat AI, Tencent Inc, China \\
  \texttt{\textrm{\{}wanglanrui,lijiangnan,linzheng,yangchenxu,wangweiping\textrm{\}}@iie.ac.cn} 
  \\
  \texttt{\textrm{\{}fandongmeng,withtomzhou\textrm{\}}@tencent.com}
  \\
  }
  
\begin{document}
\maketitle
\begin{abstract}
Empathy, which is widely used in psychological counselling, is a key trait of everyday human conversations. Equipped with commonsense knowledge, current approaches to empathetic response generation focus on capturing implicit emotion within dialogue context, where the emotions are treated as a static variable throughout the conversations. However, emotions change dynamically between utterances, which makes previous works difficult to perceive the emotion flow and predict the correct emotion of the target response, leading to inappropriate response. Furthermore, simply importing commonsense knowledge without harmonization may trigger the conflicts between knowledge and emotion, which confuse the model to choose incorrect information to guide the generation process. To address the above problems, we propose a Serial Encoding and Emotion-Knowledge interaction (SEEK) method for empathetic dialogue generation. We use a fine-grained encoding strategy which is more sensitive to the emotion dynamics (emotion flow) in the conversations to predict the emotion-intent characteristic of response. Besides, we design a novel framework to model the interaction between knowledge and emotion to generate more sensible response. Extensive experiments on \textsc{EmpatheticDialogues} demonstrate that SEEK outperforms the strong baselines in both automatic and manual evaluations.\footnote{The code is available at \url{https://github.com/wlr737/EMNLP2022-SEEK}}
\end{abstract}

\section{Introduction}

Enriching dialogue systems with human characteristics and capabilities is a hotspot in the humanlike dialogue system research area. Empathy, which is used extensively in psychological counselling \cite{OnlineMentalHealth,ESConv,CompuApproach}, is a key trait of everyday human conversations. In contrast to generating responses with controlled emotions \cite{ECM,COMAE}, the key to the empathetic dialogue system is to understand the user's emotions and generate appropriate responses. Several works concentrate on improving the empathetic models’ ability to capture contextual emotions by emotion mimicry \citep{MIME}, feedback-based adversarial generating \citep{EMPDG}, or the mixture of experts \cite{MoEL}. On the other hand, \citet{CEM,KEMP} introduce commonsense knowledge into empathetic models so as to better perceive implicit semantic information and generate more informative and empathetic response.
\begin{figure}
    \centering
    \includegraphics[width=0.50\textwidth]{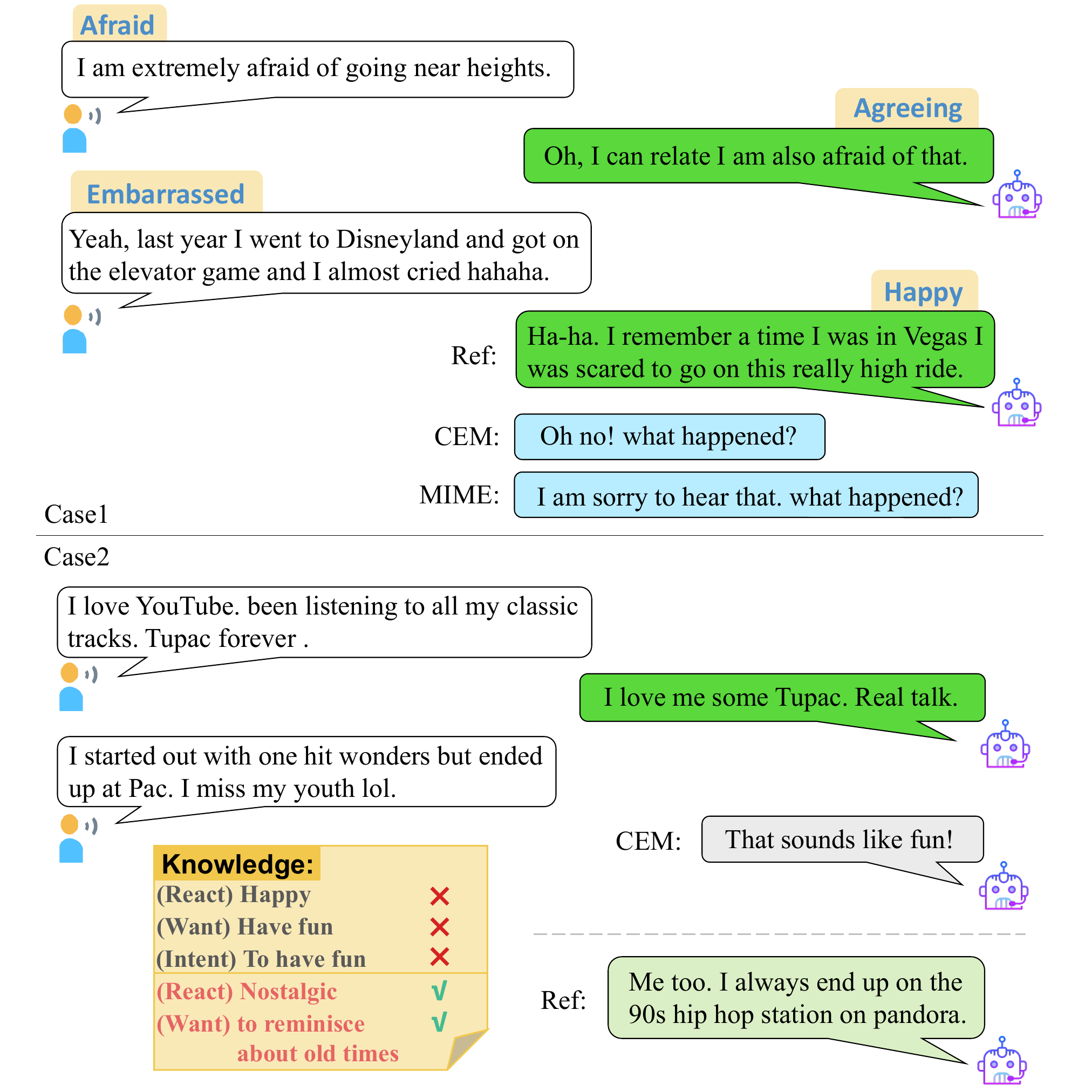}
    \caption{Two cases of multi-turn Empathetic Dialogues. The first case shows the speaker's emotion went from fear at the beginning of the conversation to an embarrassed self-deprecation, ending with a happy mood. And the second case shows that CEM chooses the wrong  knowledge leading to inappropriate response.
}
    \label{fig: cases}
\end{figure}

However, the existing works are all about the dialogue-level emotional perception \cite{MoEL,MIME,EMPDG,CEM,KEMP}. Since emotions change dynamically throughout conversations, the coarse modeling method at the dialogue level (recognizing the emotion of the whole conversation context) cannot capture the process of emotional dynamics and makes it difficult to predict response emotions. \citet{Taxonomy} have studied the shifting pattern of the utterances and drawn two graphs to show the most common emotion-intent ﬂow patterns (with a frequency $\geq$  5) throughout the ﬁrst four dialogue turns and the global exchanging trends of emotion-intent between speakers and listeners in the \textsc{EmpatheticDialogues} dataset. For instance, in the ﬁrst case illustrated in Fig. \ref{fig: cases}, the speaker’s emotion shifts from afraid at the beginning of the conversation to an embarrassed self-deprecation about previous experience of fearing heights (sharing such a funny story). Accordingly, it is much better that the dialogue agent should express the same self-deprecating sentiment like the gold response. Nevertheless, the baseline models have difficulty capturing subtle changes in the speaker's emotions and can only provide response according to the fear detected. Moreover, merely introducing knowledge without making emotionally logical choices may lead to logical conflicts between knowledge and emotion in the generated responses. As illustrated in the second case illustrated in Fig. \ref{fig: cases}, the CEM \cite{CEM} model chooses the wrong knowledge and is unable to correctly give empathetic responses with nostalgic overtones, which makes knowledge and emotion come into conflict.

To this end, we propose a Serial Encoding and Emotion-Knowledge interaction (SEEK) method for empathetic dialogue generation. To achieve a more fine-grained perception of emotional dynamics, we use an utterance-level encoding strategy which is more sensitive to the emotion flow in the conversations and able to predict the emotion characteristic of the response. We further introduce two new emotion-intent identification tasks to understand contextual emotion and predict the emotional and intentional trait of responses. For the problem of conflicts between knowledge and emotions, we also design a framework modeling the process of bi-directional interaction between them.  Extensive experimental results on the utterance-level annotated \textsc{EmpatheticDialogues} (ED) dataset \citep{Taxonomy} demonstrate that SEEK outperforms the strong baseline with both automatic and manual evaluation metrics.
Our contributions are summarized as follows:

 \begin{itemize}
	\item To the best of our knowledge, our work is the first to model the emotion flow that involves the process of emotional dynamics in the task of empathetic dialogue generation. In addition to the coarse emotion at the dialogue level, we introduce fine-grained emotions at the utterance level. 
	\item By modelling the bi-directional interactive selection process between commonsense knowledge and emotions, we have improved not only the ability to recognize contextual emotions, but also the ability to filter out unreasonable external knowledge, allowing the model to generate more sensible empathetic responses.
	\item The automatic and manual evaluation on annotated-ED dataset shows that our proposed model is superior to the strong baselines and capable of generating more diverse and sensible empathetic responses.
\end{itemize}

\section{Related Work}

In order to control the emotion of the generated response, which is one of the fundamental characteristics of daily conversation, plenty of approaches \cite{ECM,COMAE,Emocontrol1,CDL,infusing} view the target emotion as a guiding information of the models’ generator.

Contrary to controlling the emotion of the target response, the task of empathetic dialogue generation requires that the models learn a proper emotion to express empathy. Numerous researchers have attempted to improve the dialogue models’ ability to respond empathetically. \citet{EDdata} proposed a benchmark and dataset to build and evaluate empathetic dialogue generation models. \citet{MoEL} learned a precise emotion distribution of the response based on mixture of experts. \citet{MIME} split the emotions into two classes and designed a framework to mimic the target emotion in a certain class. \citet{EMPDG} utilized user feedback to build a multi-resolution adversarial training framework. In addition, \citet{Perspective} and \citet{EMP-ERT} focused on the keywords and emotion cause of dialogue history to better understand the context-level emotion and recognize feature transitions between utterances.
As well, several datasets \cite{ESConv,EDOS} of empathetic dialogue generation have been published for further research. However, most of the current approaches do not pay enough attention to the emotion flow of the conversations.

Commonsense knowledge is widely used to build dialogue systems. \citet{CARE} utilize Commonsense knowledge graph to gain candidate words for generation. \citet{CEM} adopt COMET \cite{COMET}, a pre-trained language model to generate commonsense inference for retrieving implicit information of dialogue context. In addition, \citet{KEMP} construct a graph-based framework to encode the context-knowledge graph retrieved on commonsense knowledge base. The knowledge introduced into these models might become a trigger of logical conflicts due to the absence of harmony selection.

\section{Methodology}

\begin{figure*}
    \centering
    \includegraphics[width=0.98\textwidth]{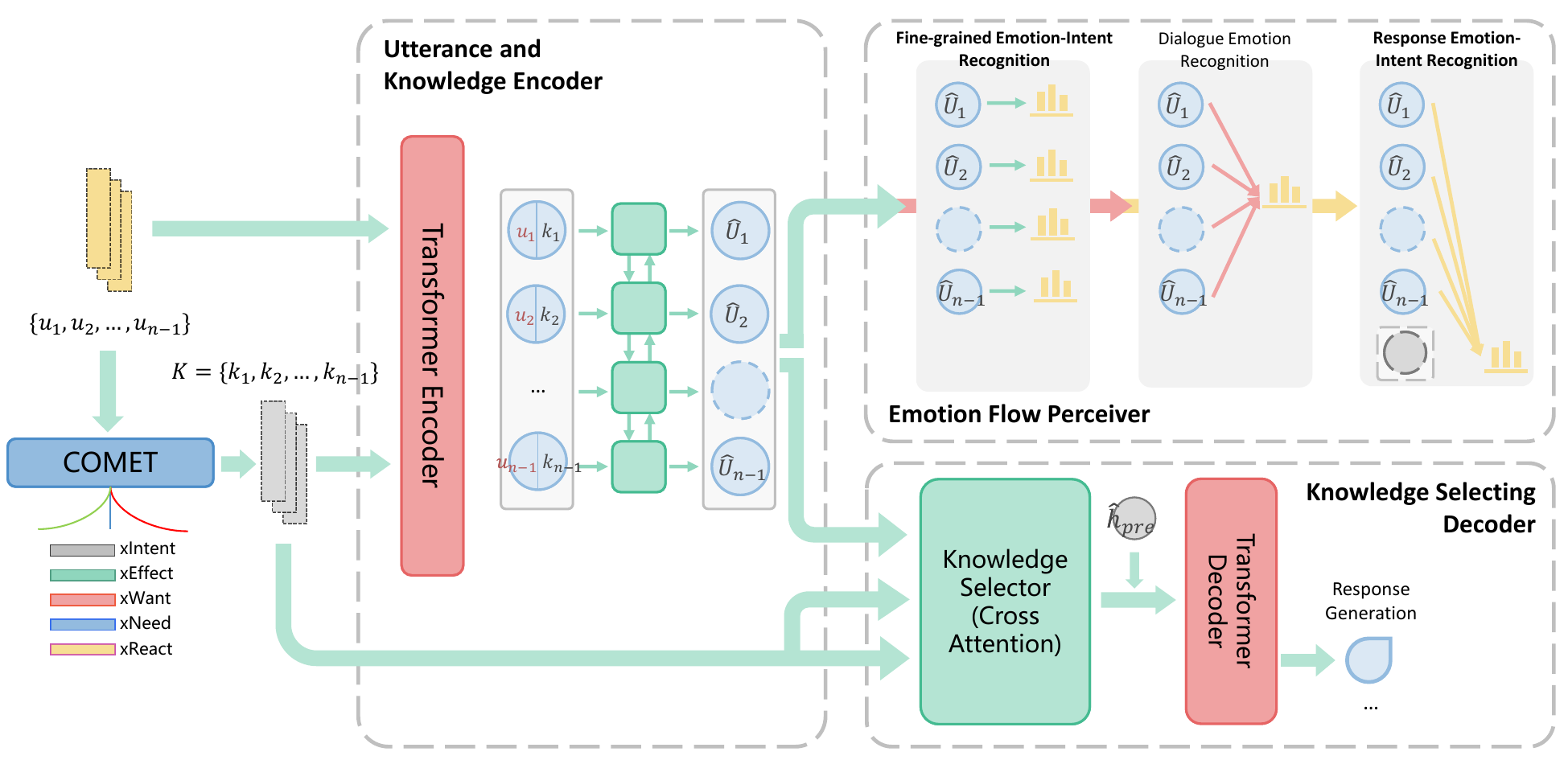}
    \caption{An overall architecture of our proposed model. }
    \label{fig: model}
\end{figure*}
\subsection{Task Formulation} \label{subsection3.1}
The task of empathetic dialogue generation is to generate empathetic responses based on the historical context. Given a dialogue $D$, where the context and the target response are denoted as $C=[C_1,...,C_{N-1}]$ and $Y$ respectively, with a emotion label of the whole context $e_c$. Additionally, a given sequence of emotion-intent labels $\boldsymbol{EI} = [\boldsymbol{ei}_1,...,\boldsymbol{ei}_{N-1},\boldsymbol{ei}_{Y}]$ of the corresponding utterances in $D$, which includes the 32 emotion categories, and 9 common intent classes. Our goal is to generate the next utterance $Y$, which is fluent and coherent to the context, and express empathy to the speaker's situation and feelings.

\subsection{Utterance and Knowledge Encoder}
 
\textbf{Utterance Encoding: }
To get a precise representation of each utterance, we firstly encode the context at the utterance level to extract the contextual information. We employ Transformer \cite{TRANSFORMER} to encode the utterance. The embedding of the input is the sum of the word embedding, positional embedding, and dialogue state embedding. Following previous work, we prepend the utterance $u_i$ with [CLS] token to obtain the utterance input $C_i = [w_{CLS},w_1, w_2 ,..., w_{L_i}]$. The embedding is then fed into the Transformer, and we obtain the representation:
\begin{equation}
\boldsymbol{H}_{U_i} =\mathbf{TRS}_{Enc}({EMB}_{C_i}), 
\label{eq: 1}
\end{equation}
where $\boldsymbol{H}_{U_i} \in {\mathbb{R}^{L_n \times d}}$, $L_n $ is the length of the utterance, and $d$ is the hidden size of the encoder.
We take the representation of [CLS] to represent the utterance:
\begin{equation}
\boldsymbol{U}_i = \boldsymbol{H}_{U_i}[0]. 
\label{eq: 2}
\end{equation}

\textbf{Knowledge Encoding: }
In order to generate high-quality commonsense inferences for the corresponding context, we utilize COMET \cite{COMET}, which is a pre-trained GPT \cite{GPT} language model and fine-tuned on ATOMIC \cite{ATOMIC}, to generate five types of commonsense knowledge: the effect of the person (xEffect), the reaction of the person speaking the corresponding sentence (xReact), the intent before the person speaking (xIntent), what the person needs (xNeed), and what the person wants after speaking the sentence (xWant). Appending these five special relation tokens after the utterance and feeding them into COMET, we get 5 commonsense inferences texts for each relation of input utterance and then concatenate them to $\boldsymbol{\mathcal{K}}_i$. Similarly, we encode the knowledge text using the same Transformer Encoder, and average the encoded hidden state via mean pooling \citep{Meanpooling}:

\begin{equation}
\begin{aligned}
&\boldsymbol{H}_{K_i} =\mathbf{TRS}_{Enc}(\boldsymbol{\mathcal{K}}_i)\\
&\boldsymbol{K}_i =\mathbf{Mean}(\boldsymbol{H}_{K_i})
\label{eq: 3}
\end{aligned}
\end{equation}


\subsection{Emotion Flow Perceiver}
Regarding the task of emotional understanding of each utterance as a tagging task, we use a Bi-LSTM to model the emotion dynamics and the interactions between different utterances for the contextual understanding process. 

The input of Bi-LSTM is the concatenation of the encoded utterances and knowledge:
\begin{equation}
\begin{aligned}
&\boldsymbol{a}_i = [\boldsymbol{U}_i ; \boldsymbol{K}_i], \\
&\boldsymbol{\hat{U}}_i = \mathbf{BiLSTM}(\boldsymbol{W}_a \boldsymbol{a}_i),
\label{eq: 4}
\end{aligned}
\end{equation}
where $\boldsymbol{W}_a \in { \mathbb{R}^{2d \times d}}$ is a trainable weight, and $\boldsymbol{\hat{U}}_i \in { \mathbb{R}^{2d}}$ represents the processed utterance representation. 
\subsubsection{Fine-grained Emotion Recognition}
For better understanding of the conversation, we pass $\boldsymbol{\hat{U}}_i$ through a tagging classifier to produce a fine-grained emotion-intent tagging distribution $P_{tag} \in { \mathbb{R}^{t}}$:

\begin{equation}
P_{tag}(\boldsymbol{ei}_i) = \mbox{Softmax}(\boldsymbol{W_e} \boldsymbol{\hat{U}}_i)
\label{eq: 5}
\end{equation}
where $t$ is the number of emotion-intent categories.

We train the tagging module with the cross-entropy loss between the predicted distribution and the ground truth label for a conversation context:
\begin{equation}
\mathcal{L}_{emo} = -\sum_{i=1}^{N-1}log(P_{tag}(\boldsymbol{ei}_i)). 
\label{eq: 6}
\end{equation}
\subsubsection{Response Emotion-Intent Prediction}

The shift in emotion and intent in empathetic dialogue conforms to an intuitive pattern.
We use the attention mechanism to learn the shift pattern of emotion and intent between utterances.
\begin{equation}
\begin{aligned}
&\mathbf{\hat{h}}_{pre}=\mbox{attention}([\boldsymbol{\hat{U}}_1, \boldsymbol{\hat{U}}_2,...,\boldsymbol{\hat{U}}_{N-1}]),  \\
&P_{pre} = \mbox{Softmax}(\boldsymbol{W_p} \boldsymbol{\mathbf{\hat{h}}}_{pre}), 
\end{aligned}
\label{eq: 7}
\end{equation}
where $\mathbf{\hat{h}}_{pre}\in { \mathbb{R}^{2d}}$ is the representation of the predicted emotion-intent characteristic of response, and $\boldsymbol{W}_p \in { \mathbb{R}^{2d \times t}}$ is the weight vector for the linear layer. $P_{pre}$ denotes the predicted distribution of the emotion-intent of the target response, $t$ is the number of emotion and intent categories.

During training,  we then minimize the cross-entropy loss between the emotion-intent distribution of the predicted response $P_{pre}$ and the ground truth label $\boldsymbol{ei}_N$ of the target response :

\begin{equation}
\mathcal{L}_{pre} = -log(P_{pre}(\boldsymbol{ei}_N)). 
\label{eq: 8}
\end{equation}
\subsubsection{Dialogue Emotion Recognition}
The sequence of utterances representation not only has the contextual information of utterances themselves but also indicates the emotional trait of the whole dialogue. Similarly, we employ the attention mechanism to summarize the holistic emotion label, based on the sequence $[\boldsymbol{\hat{U}}_1, \boldsymbol{\hat{U}}_2,...,\boldsymbol{\hat{U}}_{N-1}]$:
\begin{equation}
\begin{aligned}
&\mathbf{\hat{h}}_{dia}=\mbox{attention}([\boldsymbol{\hat{U}}_1, \boldsymbol{\hat{U}}_2,...,\boldsymbol{\hat{U}}_{N-1}]),  \\
&P_{dia} = \mbox{Softmax}(\boldsymbol{W_d} \boldsymbol{\mathbf{\hat{h}}}_{dia}), 
\end{aligned}
\label{eq: 9}
\end{equation}
where $\mathbf{h}_{dia}\in { \mathbb{R}^{2d}}$, and $\boldsymbol{W}_d \in { \mathbb{R}^{2d \times q}}$ is the weight vector for the linear layer. The $P_{dia}$ is the distribution of the dialogue emotion, $q$ is the number of available emotion categories.

The ground truth label of the dialogue emotion is denoted as $e^*$. The cross-entropy loss utilized to optimize the process of summarizing the conversational emotion is calculated by:
\begin{equation}
\mathcal{L}_{dia} = -log(P_{dia}(\boldsymbol{e}^{*})). 
\label{eq: 10}
\end{equation}

\subsection{Knowledge Selecting Decoder}

Merely introducing commonsense knowledge into empathetic models without making an emotionally logical selection to is not ideal. \citet{CEM} select commonsense inferences with an implicit procedure. On the contrary, our method models the process of bi-directional interactions between emotion and knowledge of the corresponding utterance in the conversations.

We adopt $s$ layers of Cross-Attention Transformer to perform the harmony of emotion and knowledge. Since the utterance representation sequence  $[\boldsymbol{\hat{U}}_1,\boldsymbol{\hat{U}}_2,...,\boldsymbol{\hat{U}}_{N-1}]$ passed through the three tasks of emotion, it contains emotional characteristics of the corresponding utterances. The inputs of Cross-Attention Knowledge Selector are composed of the utterance representation sequence acting as the query vector, the key and value vector which are both the knowledge text generated from the COMET model $\boldsymbol{\mathcal{K}} = [\boldsymbol{\mathcal{K}}_1,...\boldsymbol{\mathcal{K}}_{N-1}]$. The hidden representation of selected knowledge is as follows:
\begin{equation}
\boldsymbol{\mathcal{S}}=\mbox{Cross-Attention}(\boldsymbol{\hat{U}},\boldsymbol{\mathcal{K}},\boldsymbol{{\mathcal{K}}}),
\label{eq: 11}
\end{equation}
where $\boldsymbol{\mathcal{S}} \in { \mathbb{R}^{L_s \times d}}$, $L_s$ is the maximum length of the knowledge text, and $d$ is the hidden size of the model. 

Afterward,we average the harmonized knowledge via mean pooling \cite{Meanpooling}:
\begin{equation}
\boldsymbol{S}=\mbox{pooling}(\boldsymbol{\mathcal{S}}). 
\label{eq: 12}
\end{equation}

We take the Transformer Decoder as the backbone of the Decoder. We perform a concatenation operation between the averaged harmonized knowledge $\boldsymbol{S}$ and the prediction of response representation $\boldsymbol{\mathbf{\hat{h}}}_{pre}$ to get a mixture of these two types of information to represent the [SOS] token: 
\begin{equation}
\mbox{[SOS]}=\boldsymbol{W}_k([\boldsymbol{S};\boldsymbol{\mathbf{\hat{h}}}_{pre}])
\label{eq: 13}
\end{equation}
where $\boldsymbol{W}_k \in {\mathbb{R}^{2d \times d}}$ is the weight vector for the linear layer.

At the training stage, we prepend the target response $u_N=[y_1,...,y_T]$ with the $\mbox{[SOS]}$ token and get the final input of the Decoder $Y=[\mbox{[SOS]},y_1,...,y_T]$.The training loss is the standard negative log-likelihood (NLL) loss on the target response $u_N$:

\begin{equation}
\mathcal{L}_{nll} = -\sum_{t=1}^{T}log(P(y_t|C,y_{<t}). 
\label{eq: 14}
\end{equation}
\subsection{Training Objectives}

During the training process, we need to minimize three classification losses and a response generation loss. The classification losses are weighted equally:

\begin{equation}
\mathcal{L}_{cls} = \mathcal{L}_{tag} +\mathcal{L}_{pre} +\mathcal{L}_{dia}. 
\label{eq: 15}
\end{equation}

In order to improve the diversity of the generated response, we adopt Frequency-Aware Cross-Entropy (FACE) \cite{FACE} as an additional loss to penalize high-frequency tokens, similar to \citet{CEM}:
\begin{equation}
\mathcal{L}_{div} = -\sum_{t=1}^{T}\sum_{i=1}^{V}w_i\delta_t(c_i)log(P(y_t|C,y_{<t}), 
\label{eq: 16}
\end{equation}
where $w_i$ is a frequency weight value of the $i$-th token in the vocabulary $V$, $c_i$ represents a candidate token in the vocabulary and $\delta_t(c_i)$ is a function indicate whether $c_i$ equals to the ground truth token $y_t$.

Lastly, all the parameters for our proposed model are jointly trained and optimized by minimizing the weighted sum of the three mentioned losses:
\begin{equation}
\mathcal{L} = \alpha \mathcal{L}_{nll} + \beta \mathcal{L}_{cls} +\gamma \mathcal{L}_{div}, 
\label{eq: 17}
\end{equation}
where $\alpha$, $\beta$, and $\gamma$ are hyper-parameters used to balance three losses. In our experiments, we set $\alpha$= 1, $\beta$= 1, and $\gamma$= 1.5.
\section{Experimental Setup}

\begin{table*}[]
\centering
\scalebox{1.0}{
\begin{tabular}{@{}lcccccc@{}}
\toprule
\textbf{Models} &
  \textbf{PPL} &
  \textbf{Dist-1} &
  \textbf{Dist-2} &
  \textbf{\begin{tabular}[c]{@{}c@{}}DE Acc. \end{tabular}} &
  \textbf{\begin{tabular}[c]{@{}c@{}}UEI Acc. \end{tabular}} &
  \textbf{\begin{tabular}[c]{@{}c@{}}REI Acc. \end{tabular}} \\ \midrule
MIME  & 37.08 & 0.31          & 1.03          & 29.38          & -              & -     \\
EmpDG & 37.77 & 0.59          & 2.48          & 30.03          & -              & -     \\
KEMP  & \textbf{36.89} & 0.61          & 2.65          & 37.58          & -              & -     \\
CEM   & 37.03 & 0.66          & 2.99          & 36.44          & -              & -     \\ \midrule
SEEK  & 37.09 & \textbf{0.73} & \textbf{3.23} & \textbf{41.85} & \textbf{34.08} & \textbf{25.67} \\ \bottomrule
\end{tabular}
}
\caption{Automatic Evaluation results of baselines and our model. The improvement of SEEK to four strong baselines is statistically significant (paired t-tests with p-values < 0.05).}

\label{tab: auto}
\end{table*}


\begin{table*}[]
\centering
\scalebox{1}{
\begin{tabular}{@{}ccccccc@{}}
\toprule
\textbf{Models} & \textbf{PPL}   & \textbf{Dist-1} & \textbf{Dist-2} & \textbf{DE Acc. } & \textbf{UEI Acc. } & \textbf{REI Acc. } \\ \midrule
\textbf{SEEK}   & \textbf{37.09} & \textbf{0.73}   & \textbf{3.23}   & \textbf{41.85}                     & 34.08                      & 25.67                      \\
\textbf{w/o Utter}        & 37.37 & 0.70 & 3.13 & 38.9  & -              & \textbf{30.41} \\
\textbf{w/o Res}          & 37.97 & 0.63 & 2.74 & 40.82 & \textbf{50.48} & -              \\
\textbf{w/o Utter \& Res} & 38.48 & 0.60 & 2.70 & 39.7  & -              & -              \\
\textbf{w/o Emo}          & 37.67 & 0.61 & 2.66 & 41.27 & 35.88          & 23.37          \\
\textbf{w/o Know}         & 37.35 & 0.31 & 1.19 & 41.07 & 33.53          & 25.58          \\
\textbf{+ Others know}    & 37.50 & 6.90 & 2.88 & 38.25 & 34.43          & 24.32          \\
\textbf{+ Context Enc}    & 38.68 & 0.67 & 2.60 & 41.81 & 32.86          & 24.45          \\ \bottomrule
\end{tabular}
}
\caption{Ablation study of our proposed model SEEK. The best results are marked with bold.}
\label{tab: ablation}
\end{table*}

\subsection{Dataset}
Our experiments are conducted on the utterance-level annotated \textsc{EmpatheticDialogues} (ED) \cite{EDdata,Taxonomy}. ED is a large-scale multi-turn dialogue dataset that contains 25k empathetic conversations between a speaker and a listener. 
ED provides 32 evenly distributed emotion labels which are common in daily chats. However, the emotion labels of ED dataset are on the context level, there are no explicit signals for utterance-level emotions. 
\citet{Taxonomy} annotated ED dataset with 41 new categories of utterance-level emotional and intentional labels, which provide fine-grained information about the empathetic dialogues in ED dataset.
\subsection{Baselines}
We select several strong baseline models for comparison, including:\\
\textbf{MIME}: \citet{MIME} proposed a Transformer-based model employing mimicry strategy to sample the emotion of target responses based on the detected user emotion. The emotions are separated into two classes (positive and negative). The model utilizes a VAE to get the representations of the mimicking and non-mimicking emotions. \\
\textbf{EmpDG} \cite{EMPDG}: An adversarial training framework is composed of an empathetic generator and a semantic-emotional discriminator. The discriminator ensures that the responses generated by the generator are relevant to the context and also empathetic. The converged generator trained on the adversarial framework can generate empathetic responses with high diversity.\\
\textbf{KEMP}:  \citet{KEMP} employed a graph encoder to extract the contextual and concept information of the context graph constructed on external knowledge. The knowledge-enriched context graph contains emotional dependencies which helps to understand the emotion characteristic of conversations. \\
\textbf{CEM}: \citet{CEM} use COMET to generate commonsense knowledge based on the last utterance said by the speaker in dialogue. The authors use five specific prefixes (xIntent, xEffect, xWant, xNeed, xReact) to obtain five types of knowledge corresponding to the last utterance. 
The model can generate more informative empathetic responses.

\subsection{Implementation Details}
We implement our model using Pytorch \cite{pytorch}, and utilize Adam \cite{Adam} optimizer to optimize the model. We use 300-dimensional pre-trained GloVE vectors \cite{Glove} to initialize the word embeddings, which are shared between the encoder and the decoder. During the training stage, the learning rate is initialed as 0.0001 and we vary the learning rate following \citet{TRANSFORMER}. Our model is trained on one NVIDIA Geforce RTX 3090 GPU using a batch size of 32 and the early stopping strategy. For other settings, such as dropout rate, maximum decoding steps, and so forth, we keep the same as \citet{CEM}. The training time of SEEK is about 3 hours for around 27000 iterations.

\subsection{Automatic Evaluation}
Since \citet{HowNot} had proved that some automatic metrics based on word overlapping might be improper to evaluate the dialogue systems, such as BLEU \cite{BLEU} and ROUGE \cite{rouge}, we adopt Perplexity (\textbf{PPL}) and Distinct-n (\textbf{Dist}-$n$) \cite{Dist} as the main automatic metrics of generation quality. For the conversational emotion recognition and our newly introduced two tasks including fine-grained emotion-intent tagging and response emotion-intent prediction, we employ dialogue emotion accuracy (DE Acc.), utterance emotion-intent accuracy (UEI Acc.) and response emotion-intent accuracy (REI Acc.). 

To examine whether SEEK can generate more sensible response with fine-grained emotion recognition, we compare the performance of our model with the strong baselines. As shown in Table \ref{tab: auto}, the diversity scores (\textbf{Dist-1} and \textbf{Dist-2}) of SEEK outperform all of the baselines, which indicates our models can generate more informative response based on the external knowledge. We attribute this improvement to the knowledge selector and the predicted emotion of the target responses, with which the cross-attention mechanism helps to select the related knowledge based on the contextual information of utterances, and the predicted vector provides additional information of the generating process.

To prove if SEEK has better understanding of the dialogue emotion, we list the accuracy of the baselines and our proposed model. Remarkably, SEEK surpasses all of the baselines by a large margin, we attribute the increase of performance to the two fine-grained tasks we introduced. The better comprehension of the utterances in dialogue, the more accuracy it takes. In terms of the two new accuracy scores, UEI Accuracy and REI Accuracy, SEEK reaches satisfying performances, as the number of the categories of these two tasks are 41.

\begin{table}[]
\centering
\scalebox{1}{
\begin{tabular}{@{}llclclc@{}}
\toprule
\textbf{Models} &  & \textbf{Coh.} &           & \textbf{Emp.} &           & \textbf{Flu.} \\ \midrule
\textbf{MIME}   &  & 2.84          &           & 2.97          &           & 2.87          \\
\textbf{EmpDG}  &  & 2.85          &           & 2.78          &           & 2.76          \\
\textbf{KEMP}   &  & 2.73          &           & 2.80           &           & 2.80           \\
\textbf{CEM}    &  & 2.82          &           & 2.99          &           & 2.75          \\ \midrule
\textbf{SEEK}   &  & \textbf{2.91} & \textbf{} & \textbf{3.02} & \textbf{} & \textbf{3.07} \\ \bottomrule
\end{tabular}
}
\caption{Human evaluation results. We apply Fleiss's Kappa, denoted as $\kappa$, to measure inter-annotator agreement, where $ 0.4 < \kappa < 0.6 $ indicates moderate agreement.}
\label{tab: human1}
\end{table}

\subsection{Human Evaluation}
Following previous works, we conduct a human evaluation based on three aspects: \textit{coherence} (\textbf{Coh.}): How much does the response relevant to the context? \textit{empathy} (\textbf{Emp.}): How much does the model know about the speaker's situation and emotion characteristic? Does the model respond empathetically enough or give suggestions? \textit{fluency} (\textbf{Flu.}): How much the generated response obey the grammar? We randomly choose 100 dialogues and assign the responses generated by the models to three crowd-sourced workers for the evaluation. Each aspect is on a scale of 1 to 5. Moreover, considering the variation between different individuals, we conduct another human A/B test to directly compare our method with other baselines. Three professional annotators score the questionnaire of the response pairs to choose one of the responses in random order or select "Tie" when the quality of provided sentence is difficult to distinguish. As the results of the human rating and A/B test are shown in Table \ref{tab: human1} and table \ref{tab: human2}, SEEK outperforms the baselines in all the three aspects.

\begin{table}[]
\scalebox{0.9}{
\begin{tabular}{ccccc}
\toprule
\textbf{Comparisons}    & \textbf{Aspects} & \textbf{Win}  & \textbf{Lose} & \textbf{Tie}  \\ \hline
               & Coh.    & \textbf{24.3} & 17.1 & 58.6 \\
SEEK vs. MIME  & Emp.    & \textbf{31.4} & 22.2 & 46.4 \\
               & Flu.    & \textbf{28.6} & 25.9 & 45.5 \\ \hline
               & Coh.    & \textbf{32.1} & 26.3 & 41.6 \\
SEEK vs. EmpDG & Emp.    & \textbf{35.5} & 27.4 & 37.1 \\
               & Flu.    & \textbf{26.9} & 22.3 & 50.8 \\ \hline
               & Coh.    & \textbf{29.2} & 25.2 & 45.6 \\
SEEK vs. KEMP  & Emp.    & \textbf{28.8} & 19.9 & 51.3 \\
               & Flu.    & \textbf{38.7} & 15.6 & 45.7 \\ \hline
               & Coh.    & \textbf{27.3} & 24.8 & 47.9 \\
SEEK vs. CEM   & Emp.    & \textbf{33.4} & 27.5 & 39.1 \\
               & Flu.    & \textbf{35.7} & 21.6 & 42.7 \\ \bottomrule
\end{tabular}
}
\caption{Human A/B test ($\%$) on the three aspects: coherence, empathy, and fluency. The comparison results directly shows SEEK outperforms the four baselines models.}
\label{tab: human2}
\end{table}

\begin{table*}[]
\centering
\scalebox{0.72}{
\begin{tabular}{@{}m{2.8cm}m{16cm}@{}}
\toprule
\textbf{}                & \textit{Speaker:} I love YouTube. I've been listening to all my \textbf{classic} tracks. Tupac forever. (\textbf{Nostalgic}) \\ \cmidrule(l){2-2} 
\textbf{Context}         & \textit{Listener:} I love me some Tupac. Real talk.   (\textbf{Acknowledging})                                         \\ \cmidrule(l){2-2} 
\textbf{}                & \textit{Speaker:} I started out with \textit{One Hit Wonders} but ended up at Pac. \textbf{I miss my youth lol}.  (\textbf{Nostalgic})\\ \midrule
\textbf{Speaker Emotion} & Nostalgic                                                                             \\ \midrule
\textbf{MIME}            & I am sure you will be able to get a new one.                                          \\
\textbf{EmpDG}           & I am sure you will be fine.                                                           \\
\textbf{KEMP}            & I am sure you will be fine.                                                           \\
\textbf{CEM}             & That sounds like fun!                                                                 \\ \midrule
\textbf{SEEK}            & I do not like that, but i think they will be able to {\color[HTML]{FE0000} get the old times} .    (\textbf{Nostalgic})            \\ \midrule
\textbf{Golden}          & Me too. I always end up on the 90s hip hop station on Pandora.  (\textbf{Nostalgic})                        \\ \bottomrule \\ \toprule
\textbf{} &
  \begin{tabular}[c]{@{}l@{}}\textit{Speaker:} Yeah about 10 years ago I had a horrifying experience. It was 100\% their fault, but they hit the \\ water barrels and survived. They had no injuries, but they almost ran me off the road. (\textbf{Guilty}) \end{tabular} \\ \cmidrule(l){2-2} 
\textbf{Context}         & \textit{Listener:} Did you suffer any injuries?     (\textbf{Questioning})                                           \\ \cmidrule(l){2-2} 
\textbf{} &
  \textit{Speaker:} No, I was not hit. It turned out they were drunk. I felt guilty, but realized it was his fault. (\textbf{Guilty}) \\ \midrule
\textbf{Speaker Emotion} & Guilty                                                                                \\ \midrule
\textbf{MIME}            & I would be too. I would be so scared.                                                 \\
\textbf{EmpDG}           & I am sorry to hear that.                                                              \\
\textbf{KEMP}            & I am sorry to hear that.                                                              \\
\textbf{CEM}             & That is so sad.                                                                       \\ \midrule
\textbf{SEEK}            & That is pretty scary! I am {\color[HTML]{FE0000} glad} that you were able to {\color[HTML]{FE0000} get in out}!  (\textbf{Sympathizing})        \\ \midrule
\textbf{Golden}          & Why did you feel guilty? People really shouldn't drive drunk.    (\textbf{Questioning})  \\ \bottomrule
\end{tabular}
}
\caption{Two cases of generated responses by SEEK and the baselines. We annotated each turn with the emotional or intentional labels at the end of the utterances. The words relevant to the predicted labels in SEEK's response are highlighted in red. }
\label{tab: casestudy}
\end{table*}


\subsection{Ablation Studies}



To study the effect of tasks and modules employed in our model, we remove the newly introduced tasks and the interaction process between emotion and knowledge. Additionally, we replace the knowledge type and encoding strategy respectively. The results are demonstrated in Table \ref{tab: ablation}.

Removing the task of fine-grained Utterance Emotion-Intent tagging and Response Emotion-Intent prediction (\textbf{w/o Utter}, \textbf{w/o Res}, and \textbf{w/o Utter \& Res}) causes the drop of accuracy of dialogue emotion recognition and generative quality, as these variants lose the fine-grained understanding of the dialogue and the ability to predict the emotion-intent characteristics of the target response. 

The margin between the variant (\textbf{w/o Emo}) without emotional harmonization of the knowledge and SEEK proves the importance of the interaction between knowledge and emotion-intent from the Knowledge Selection module of our model. The variant without knowledge (\textbf{w/o Know}) indicates the importance of external knowledge for the diversity of responses the model generated.

Moreover, the decreased performance by replacing the type of knowledge \textbf{+ Others Know} and the encoding strategy \textbf{+ Context Enc} shows the superiority of our method. Using \textit{Others} type of knowledge in our model rather than \textit{PersonX} results in a considerable decrease in all performance, which indicates that the PersonX type of commonsense helps the model to understand the utterances more effectively. The encoding strategy employed by baselines (as the variant \textbf{+ Context Enc} used) emphasizes on overall understanding of the whole conversation, ignoring an accurate grasp of utterances, which leads to a decline of performance.

Remarkably, the UEI Accuracy of \textbf{w/o Utter} and REI Accuracy of \textbf{w/o Res} are higher than SEEK. This is possibly due to the noise of the utterance label of annotated ED dataset and the subtle differences between intent categories (e.g. agreeing and acknowledging, counselling and questioning), which means the classification supervision signal of utterances or the response will make the input vector of attention module harder and lose some information of other classes. The loss of information about the hidden states may confuse another classifier and leads to a decrease in accuracy. In any case, although there exists a trade-off between these two tasks, they can simultaneously improve the ability of the model to generate more sensible empathetic responses by modeling the emotion flow.

\subsection{Case Study}
The first case of figure \ref{fig: cases} illustrates how emotion shifts during a multi-turn conversation. For better compares generated responses of our model and the baselines, we show two of the generated result of our model and baselines in  Table \ref{tab: casestudy}. In the first case, the baselines failed to give responses with nostalgic overtones, similar to the commonsense knowledge demonstrated in figure \ref{fig: cases}, where CEM choose the wrong knowledge to generate response with a \textit{happy} emotion and the intent \textit{to have fun}. On the contrary, SEEK successfully gives a response with more sensitive and accurate emotional perception. Similarly, in the second case, all of the baselines generate responses 
based on the explicit emotion \textit{guilty}, without fine-grained understanding which is more accurate. Unlike the baselines, SEEK respond sensitively with sympathizing intent.

We further draw a heat map to illustrate the cross-attention weights of commonsense knowledge in a certain case. The detailed information of that case and analysis will be shown in Appendix \ref{sec:appendix}.

\section{Conclusion}
In this paper, we study the task of empathetic dialogue generation. The strong baselines ignore emotion flow of the conversations. We therefore proposed a Serial Encoding and Emotion-Knowledge interaction (SEEK) method for empathetic dialogue generation, to predict the correct emotion of the target response by perceiving the emotion flow of the context and harmonizing commonsense knowledge with fine-grained emotions to avoid conflicts. Experiments on the utterance-level annotated \textsc{EmpatheticDialogues} show that our model outperforms the baselines, and the ablation studies indicate that all the components of our model, the encoding strategy, and the commonsense knowledge work. 

In the future, we will focus on further usage (e.g. providing online-emotion aid) of empathetic systems and try to improve normalization capabilities of our model on other datasets.


\section*{Limitations}
The limitation of our work mainly comes from the shortage of datasets in the task of empathetic dialogue generation. Although there are several newly released large-scale datasets \citep{ESConv, EDOS}, most of the research can only be carried out on the English corpus \textsc{EmpatheticDialogues}. Another limitation is the problem of evaluation metrics. As mentioned in \citet{HowNot}, the scores of standard automatic evaluation metrics are not consistent with human evaluation results. The lack of task-specifically automatic metrics makes it troublesome for evaluating empathetic dialogue generation.

\section*{Ethical Considerations}

The data \citep{EDdata,Taxonomy} used in our work is all drawn from open-source datasets. The conversations of the dataset are around given emotions and carried out by employed crowd-sourced workers, with no personal privacy issues involved.

\section*{Acknowledgments}

This work was supported by National Natural Science Foundation of China (No. 61976207, No. 61906187).

\bibliography{anthology}
\newpage

\appendix

\section{More Cases} \label{sec:appendix}
To show the process of knowledge selection of our proposed model, we clearly show the attention weights on the commonsense knowledge in Table \ref{tab: case2}. We firstly get the weights matrix from Cross-Attention outputs and search the words in the knowledge text by the index of high-value elements. To directly show the selecting process, we mark the knowledge words based on the color in the heat map we drew: the higher weight the knowledge words have the darker blue marks them in the table.

In this case, the context of the case is mainly about a couple of parents asking for the gender of the baby in a hospital and the COMET totally model generates 25 commonsense inferences based on it. The speaker reacts excitedly to knowing the gender of their baby which infers something to celebrate, and SEEK chooses the correct knowledge and expresses congratulation.
\begin{table*}[]
\centering
\scalebox{0.8}{
\begin{tabular}{llllll}
\hline
Type &
  \multicolumn{1}{l|}{x\_intent} &
  \multicolumn{1}{l|}{x\_need} &
  \multicolumn{1}{l|}{x\_want} &
  \multicolumn{1}{l|}{x\_effect} &
  x\_react \\ \hline
 &
  \multicolumn{1}{l|}{\colorbox{blue!5}{to see the \colorbox{blue!60}{baby}}} &
  \multicolumn{1}{l|}{\colorbox{blue!5}{to have an ultrasound}} &
  \multicolumn{1}{l|}{\colorbox{blue!5}{to see what the \colorbox{blue!60}{baby} is}} &
  \multicolumn{1}{l|}{\colorbox{blue!5}{to see the \colorbox{blue!60}{baby}}} &
  {\colorbox{blue!45}{happy}} \\ 
 &
  \multicolumn{1}{l|}{\colorbox{blue!5}{to know the gender}} &
  \multicolumn{1}{l|}{\colorbox{blue!5}{to see the ultrasound}} &
  \multicolumn{1}{l|}{\colorbox{blue!5}{to show it to their friends}} &
  \multicolumn{1}{l|}{\colorbox{blue!5}{to see the \colorbox{blue!50}{gender}}} &
  {\colorbox{blue!5}{excited}} \\ 
Knowledge &
  \multicolumn{1}{l|}{\colorbox{blue!5}{to know the sex}} &
  \multicolumn{1}{l|}{\colorbox{blue!5}{to have the ultrasound}} &
  \multicolumn{1}{l|}{\colorbox{blue!5}{to show it to everyone}} &
  \multicolumn{1}{l|}{\colorbox{blue!5}{to see the ultrasound}} &
  {\colorbox{blue!5}{surprised}} \\ 
 &
  \multicolumn{1}{l|}{\colorbox{blue!5}{to be informed}} &
  \multicolumn{1}{l|}{\colorbox{blue!5}{to have a \colorbox{blue!60}{baby}}} &
  \multicolumn{1}{l|}{\colorbox{blue!5}{to show it to others}} &
  \multicolumn{1}{l|}{\colorbox{blue!5}{to be happy}} &
  {\colorbox{blue!5}{joyful}} \\
 &
  \multicolumn{1}{l|}{\colorbox{blue!5}{none}} &
  \multicolumn{1}{l|}{\colorbox{blue!5}{to get the ultrasound}} &
  \multicolumn{1}{l|}{\colorbox{blue!5}{to see the \colorbox{blue!60}{baby}}} &
  \multicolumn{1}{l|}{\colorbox{blue!5}{we get excited}} &
  {\colorbox{blue!5}{relieved}} \\ \cline{2-6} 
Context &
  \multicolumn{5}{l}{\begin{tabular}[c]{@{}l@{}}We asked the doc to put the ultrasound in an envelope so we could record our reaction to the gender reveal. \\ I was very happy when I finally saw it! (\textbf{Excited})\end{tabular}} \\ \hline
SEEK &
  \multicolumn{5}{l}{\textbf{Congratulations!}} \\ \hline
Gold &
  \multicolumn{5}{l}{\textbf{Congrats!} what gender did your child end up being?} \\ \hline
\end{tabular}

}
\caption{The visualization of the cross-attention weights of selecting knowledge in SEEK.}
\label{tab: case2}
\end{table*}

\end{document}